%% file: main.tex
\pdfoutput=1

\documentclass[11pt]{article}

\usepackage{ACL2023}
\usepackage{times}
\usepackage{latexsym}

\usepackage[T1]{fontenc}

\usepackage[utf8]{inputenc}

\usepackage{microtype}

\usepackage{inconsolata}
\usepackage{url}

\input{math_commands.tex}

\usepackage{amsmath}
\usepackage{multirow}
\usepackage{xcolor}

\usepackage{xcolor,colortbl}
\definecolor{lightblue}{rgb}{0.68, 0.85, 0.9}
\definecolor{lavender}{rgb}{0.9, 0.9, 0.98}
\definecolor{lightyellow}{rgb}{1.0, 1.0, 0.88}
\definecolor{magicmint}{rgb}{0.67, 0.94, 0.82}
\definecolor{palepink}{rgb}{0.98, 0.85, 0.87}
\definecolor{bubbles}{rgb}{0.91, 1.0, 1.0}

\usepackage{microtype}
\usepackage{graphicx}
\usepackage{subfigure}
\usepackage{amssymb}
\usepackage{CJKutf8}
\usepackage{algorithm}
\usepackage{algorithmic}
\usepackage{enumitem}
\algsetup{linenosize=\small}
\usepackage{wrapfig}
\usepackage{cleveref}
\usepackage{multicol}
\usepackage{booktabs}
\usepackage{xspace}
\usepackage{adjustbox}
\usepackage{subfiles}
\usepackage{latexsym}
\usepackage{hyperref}

\newcolumntype{P}[1]{>{\centering\arraybackslash}p{#1}}
\usepackage{subfigure}

\usepackage{placeins}
\usepackage[french,vietnamese,main=english]{babel}
\usepackage{tikz}
\babelprovide[import]{thai}

\newcommand*{\affaddr}[1]{#1} 

\newcommand*{\email}[1]{\textrm{#1}}

\title{Panda LLM: Training Data and Evaluation for Open-Sourced Chinese Instruction-Following Large Language Models}

\author{Fangkai Jiao\thanks{\; Equal contribution, order decided by coin flip.} ~~Bosheng Ding\footnotemark[1]~~\thanks{\; Corresponding Author.}~~ \textbf{Tianze Luo\footnotemark[1] ~~Zhanfeng Mo\footnotemark[1]}\\
\affaddr{Nanyang Technological University, Singapore}
\\
\email{\small{jiaofangkai@hotmail.com} \quad \small{\{bosheng001, tianze001, zhanfeng001\}@ntu.edu.sg}}\\}

\begin{document}
\maketitle
\begin{abstract}
  This project focuses on enhancing open-source large language models through instruction-tuning and providing comprehensive evaluations of their performance. We explore how various training data factors, such as quantity, quality, and linguistic distribution, influence the performance of instruction-tuned models trained on publicly accessible high-quality instruction datasets for both English and Chinese languages. Our goal is to supplement evaluation with quantitative analyses, providing valuable insights for the continued advancement of open-source chat models. Our model, data, and code are publicly available\footnote{\href{https://github.com/dandelionsllm/pandallm/}{https://github.com/dandelionsllm/pandallm/}} for others to use and build upon.
\end{abstract}

\input{contents/intro}
\input{contents/data}
\input{contents/train}
\input{contents/experiments}

\input{contents/future}

\section*{Acknowledgments}
We are very grateful for the support from a few large organizations, which have provided us with a large number of GPUs to support our model training. The high-performance computing power of these GPUs has provided us with strong support in the research and development of the Panda model.

\bibliographystyle{acl_natbib}
\bibliography{reference}

\end{document}

%% file: math_commands.tex
\usepackage{amsmath,amsfonts,bm}

\def\eqref#1{equation~\ref{#1}}

\def\1{\bm{1}}

\DeclareMathAlphabet{\mathsfit}{\encodingdefault}{\sfdefault}{m}{sl}
\SetMathAlphabet{\mathsfit}{bold}{\encodingdefault}{\sfdefault}{bx}{n}



%% file: contents/intro.tex
\section{Introduction}
Over the last six months, there has been a significant surge in the development and advancement of instruction-following Large Language Models (LLM) such as GPT-4 \cite{openai2023gpt}, GPT-3.5 (text-davinci-003)\footnote{\href{https://platform.openai.com/docs/models/gpt-3-5}{https://platform.openai.com/docs/models/gpt-3-5}}, ChatGPT\footnote{\href{https://chat.openai.com/}{https://chat.openai.com/}}, Claude\footnote{\href{https://www.anthropic.com/index/introducing-claude}{https://www.anthropic.com/index/introducing-claude}}, and Bard\footnote{\href{https://bard.google.com/}{https://bard.google.com/}}. These models have gained widespread popularity due to their exceptional versatility in various natural language proccessing tasks such as code writing and article editing, making them ubiquitous in various industries and significantly enhancing people's productivity \cite{ding2022gpt,zhao2023retrieving}. However, there are limitations to current off-the-shelf instruction-following large language models, including the lack of trustworthiness in generated results, lack of transparency in the model used which raises concerns about data security, and the unknown training recipe, making it difficult to customize a self-used model for specific purposes \cite{LLaMA}.

We believe that the cultivation of a strong and versatile open-source community for the development of trustable, transparent, and customizable large language models in all languages worldwide is considered the best approach to address the current issues and make the power of large language models accessible to everyone. In line with this objective, the Dandelion Project is proposed to deploy large language models that are not only accurate but also transparent, trustworthy, and customizable. The project aims to promote more accessible and inclusive AI technology that can benefit individuals regardless of their cultural differences, geographical locations, or language barriers. Through open-source access to high-quality large language models, the Dandelion Project aims to empower developers, researchers, and organizations to leverage AI's potential in various applications such as translation, chatbots, content generation, and more.



\begin{figure*}[t!]
    \centering
    \includegraphics[scale=0.7]{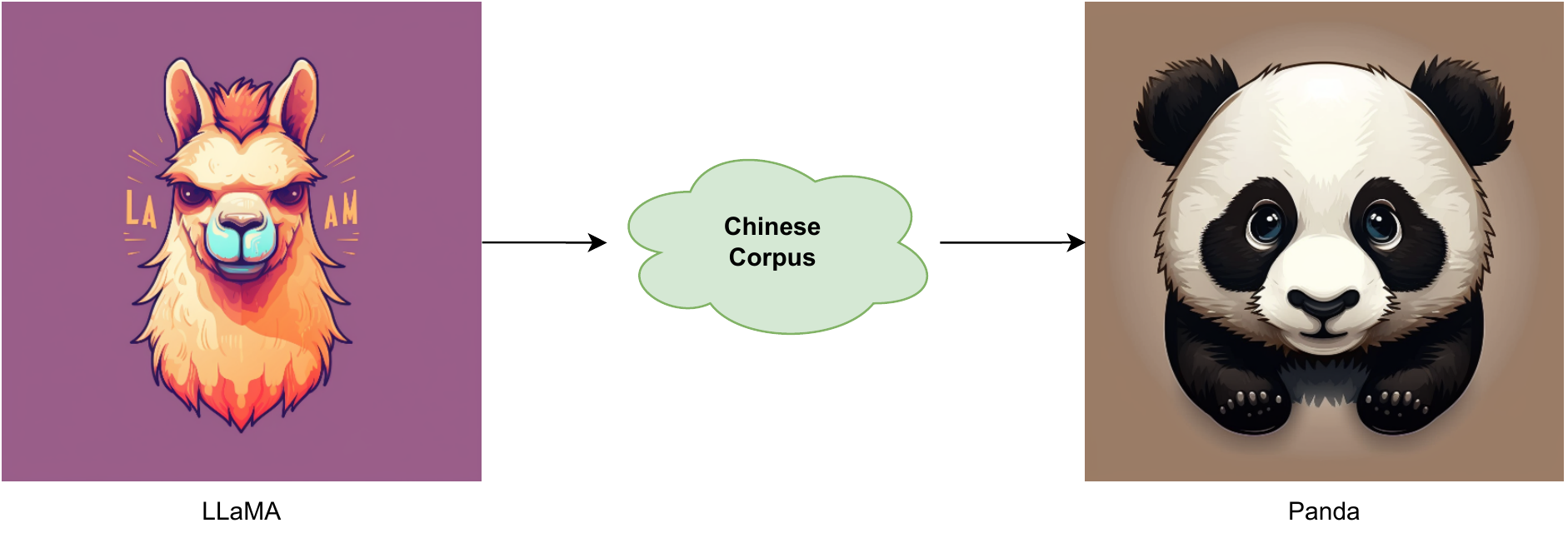}
    \caption{Illustrations of our proposed method.}
    \label{fig:method} 
\end{figure*}

This report presents the Panda LLM, which is the first open-sourced Chinese instruction-following large language model for overseas audiences. It is also the first released LLM of the Dandelion Project. Our Panda LLM model has been trained on Chinese-Wiki-2019, Chinese-News-2016, Chinese-Baike-2018, Chinese-Webtext-2019 and Translation-2019 \cite{wiki_zh} and COIG datasets \cite{COIG} with instruction-tuning \cite{instruct_tune} based on the LLaMA model \cite{LLaMA}. Anticipated future releases include progressively larger models such as Panda-13B and Panda-33B, with expected release dates in the near future.

Due to the presence of the LLaMA weight License, we can not directly publish the complete weights of the checkpoints of our Panda LLM. Therefore, we have released the difference between the parameters of the trained model and the original LLaMA weights to ensure that users with access to the LLaMA weights can still utilize these models. A script has been provided to facilitate the conversion process. To this end, the contribution of this project is three-fold:
\begin{itemize}
    \item We adopted a two-stage training approach which yielded exemplary results, surpassing all previously available open-sourced Chinese large language models with an equivalent amount of parameters (Section \ref{sec:train}).
    \item We conducted the first-ever comparative evaluation of various open-sourced Chinese large language models (Section \ref{sec:eval}).
    \item We have made available a collection of model checkpoints and the corresponding source codes, with the objective of promoting the democratization of Artificial Intelligence. These resources are intended to be of benefit not only to the academic community but also to individuals and \textbf{S}mall and \textbf{M}edium-sized \textbf{E}nterprises (SMEs).
\end{itemize}

%% file: contents/train.tex
\section{Training Receipt}
\label{sec:train}

To create a high-quality instruction-following Chinese language model under academic budget constraints, two key components are required: a robust pre-trained language model and a high-quality instruction-following dataset. In this section, we will demonstrate our process of developing the Panda LLM. We started with the powerful LLaMA base model as our foundation and further optimized its performance through fine-tuning with instruction-tuning techniques on six Chinese corpora, enabling it to perform well on a diverse range of tasks.

\subsection{Base model}
Our Panda LLM is established based on various LLaMA (Large Language Model Meta AI) models \cite{LLaMA}, including Meta's recently released LLaMA-7B, LLaMA-13B, LLaMA-33B, and LLaMA-65B, as our base models. LLaMA models, although smaller than giant commercial models like ChatGPT and GPT4, are highly performant and open-sourced, providing greater accessibility to foundation large language models across various domains with far less computing power and resources. Similar to other large language models, LLaMA works by taking a sequence of words as an input and predicts the next word to recursively generate text. 

Following recent work on large language models,
our network is based on the transformer architecture \cite{attention}. Various improvement is leveraged to enhance the model capacity, including pre-normalization \cite{rms_norm}, SwiGLU activation function and rotary embeddings \cite{rotary_emb}. As shown in Table \ref{tab:dastaset}, LLaMA models are trained on a mixture of 7 publicly available datasets, comprising of 1.4T tokens. The training configurations and model hyperparameters are shown in Table \ref{LlaMA:train}.

\begin{table*}
\begin{tabular}{ccccccc}
\hline 
LLaMA & \multicolumn{6}{c}{ Model hyper parameters } \\
\hline 
Number of parameters & dimension & \# heads & \# layers & Learn rate & Batch size & $\mathrm{n}$ tokens \\
\hline
\hline
7B & $4096$ & $32$ & $32$ & $3.0 \times 10^{-4}$ & $4 \mathrm{M}$ & $1 \mathrm{~T}$ \\
\hline 
13B & $5120$ & $40$ & $40$ & $3.0 \times 10^{-4}$ & $4 \mathrm{M}$ & $1 \mathrm{~T}$ \\
\hline 
33B & $6656$ & $52$ & $60$ & $1.5 \times 10^{-4}$ & $4 \mathrm{M}$ & $1.4 \mathrm{~T}$ \\
\hline 
65B & $8192$ & $64$ & $80$ & $1.5 \times 10^{-4}$ & $4 \mathrm{M}$ & $1.4 \mathrm{~T}$ \\
\hline
\end{tabular}
\caption{The training configuration and model hyperparameters of LLaMA models.}
\label{LlaMA:train}
\end{table*}

\subsection{Training datasets}

While many existing open-sourced large language models have demonstrated impressive performance on English language tasks, they are primarily pre-trained on English datasets, limiting their ability to understand Chinese language corpus. In this section, we address the challenge of the scarcity of high-quality Chinese instruction-following datasets in the training receipts of existing open-source LLMs. To enable our Panda LLM to acquire strong performance on Chinese datasets, we utilized the powerful instruction-tuning technique to train the base LLaMA model on a mixture of five open-sourced Chinese datasets \cite{wiki_zh}. These datasets, as shown in Table \ref{panda:dataset}, consist of 15.3 million machine comprehension samples from various language domains, such as news articles, community question-answering and translation, etc.

\begin{table*}[t!]
\centering
\begin{tabular}{cc}
\hline 
Dataset  & Ingredient  \\
\hline 
\hline
Chinese-Wiki-2019 &  1M Chinese short paragraphs.\\
Chinese-News-2016 & 2.5M Chinese news from 2014 to 2016.\\
Chinese-Baike-2018 & 1.5M Chinese QA data samples.\\
Chinese-Webtext-2019 & 4.1M Chinese high-quality QA data samples for various domains.\\
Translation-2019 & 5.2M Chinese-English translation data samples.\\
\hline
\end{tabular}
\caption{The NLP Chinese Corpus datasets for Panda LLM.}
\label{panda:dataset}
\end{table*}


Particularly, for dataset other than Chinese-Wiki-2019 and Chinese-News-2016, our model is optimized following the conditional text generation paradigm \cite{cond_gen}, in which the loss is solely calculated based on the output part, and the instruction and input parts are ignored. A fixed prompt template is utilized for the instruction across these datasets. 

After making several initial attempts to directly train models on a mixture of these datasets, we realized that our model's instruction-following performance was limited. We speculate that this is due to the insufficient number of instruction-following samples in the entire training corpus, which results in suboptimal training of our model for instruction-following tasks. 

To enhance the instruction-following capability of Panda LLM, we further incorporate the Chinese Open Instruction Generalist (COIG) dataset \cite{COIG} into our corpus. COIG is an open-sourced Chinese corpora that contains instruction-following samples from various domains, including a manually verified translated general instruction corpus, a manually annotated exam instruction corpus, a human value alignment instruction corpus, a multi-round counterfactual correction chat corpus, and a leetcode instruction corpus.  As we shall see later, extra optimization on COIG brings Panda LLM noticeable performance boost. And our model is further improved via up-sampling techniques on the COIG dataset.

\begin{table*}
\centering

\begin{tabular}{llccr}
\toprule

& Dataset & Sampling prop. & Epochs & Disk size \\
\hline 
\hline
\multirow{7}{*}{LLaMA}
& CommonCrawl & $67.0 \%$ & 1.10 & $3.3 \mathrm{~TB}$ \\
& C4 & $15.0 \%$ & 1.06 & $783 \mathrm{~GB}$ \\
& Github & $4.5 \%$ & 0.64 & $328 \mathrm{~GB}$ \\
& Wikipedia & $4.5 \%$ & 2.45 & $83 \mathrm{~GB}$ \\
& Books & $4.5 \%$ & 2.23 & $85 \mathrm{~GB}$ \\
& ArXiv & $2.5 \%$ & 1.06 & $92 \mathrm{~GB}$ \\
& StackExchange & $2.0 \%$ & 1.03 & $78 \mathrm{~GB}$ \\
\midrule
\multirow{6}{*}{Panda (ours)}
& Chinese-Wiki-2019  & 9.4\%& 1 & 1.6GB\\
& Chinese-News-2016  & 52.6\%& 1 & 9GB\\
& Chinese-Baike-2018  & 5.8\% & 1 & 1GB\\
& Chinese-Webtext-2019  & 21.6\% & 1 & 3.7GB\\
& Translation-2019 & 6.4\% & 1 & 1.1GB\\
& COIG & 4.2\%& 2 & 350MB \\

\bottomrule

\end{tabular}
\caption{Training data comparison for LLaMA and Panda. For each subset we list the sampling proportion, number of epochs, and disk size.}

\label{tab:dastaset}
\end{table*}



\subsection{Training infrastructure}
Our Panda-7B and Panda-13B models were trained on two AWS computation nodes that were equipped with 16 NVIDIA A100-80G GPUs. We leverage the standard Stochastic Gradient Descent (SGD) \cite{sgd} optimizer to train our Panda LLMs. For the Panda-7B and Panda-13B models, we set the batch sizes after gradient accumulation to $8192$ and $4096$, and the learning rates to 1e-5 with $1\%$ of the total training steps allocated for learning rate warm-up steps \cite{lr_warmup}. We disabled the weight decay for both models. During instruction tuning for the 7B model, we utilized batch sizes of 3e-5 and $128$. To facilitate efficient model training, we employed DeepSpeed\footnote{\href{https://www.deepspeed.ai/}{www.deepspeed.ai}} ZERO-1 \cite{ZERO} with \texttt{bfloat16} and gradient checkpointing. The training process took approximately 7 and 14 days for the Panda-7B and Panda-13B models, respectively. Additional training details can be found in the \texttt{config} files in our GitHub repository\footnote{\href{https://github.com/dandelionsllm/pandallm/tree/main/conf/llama/zh}{github.com/dandelionsllm/pandallm/tree/main/conf/llama/zh.}}.


\begin{figure}[t!]
    \centering
 \subfigure{\includegraphics[width=0.45\textwidth]{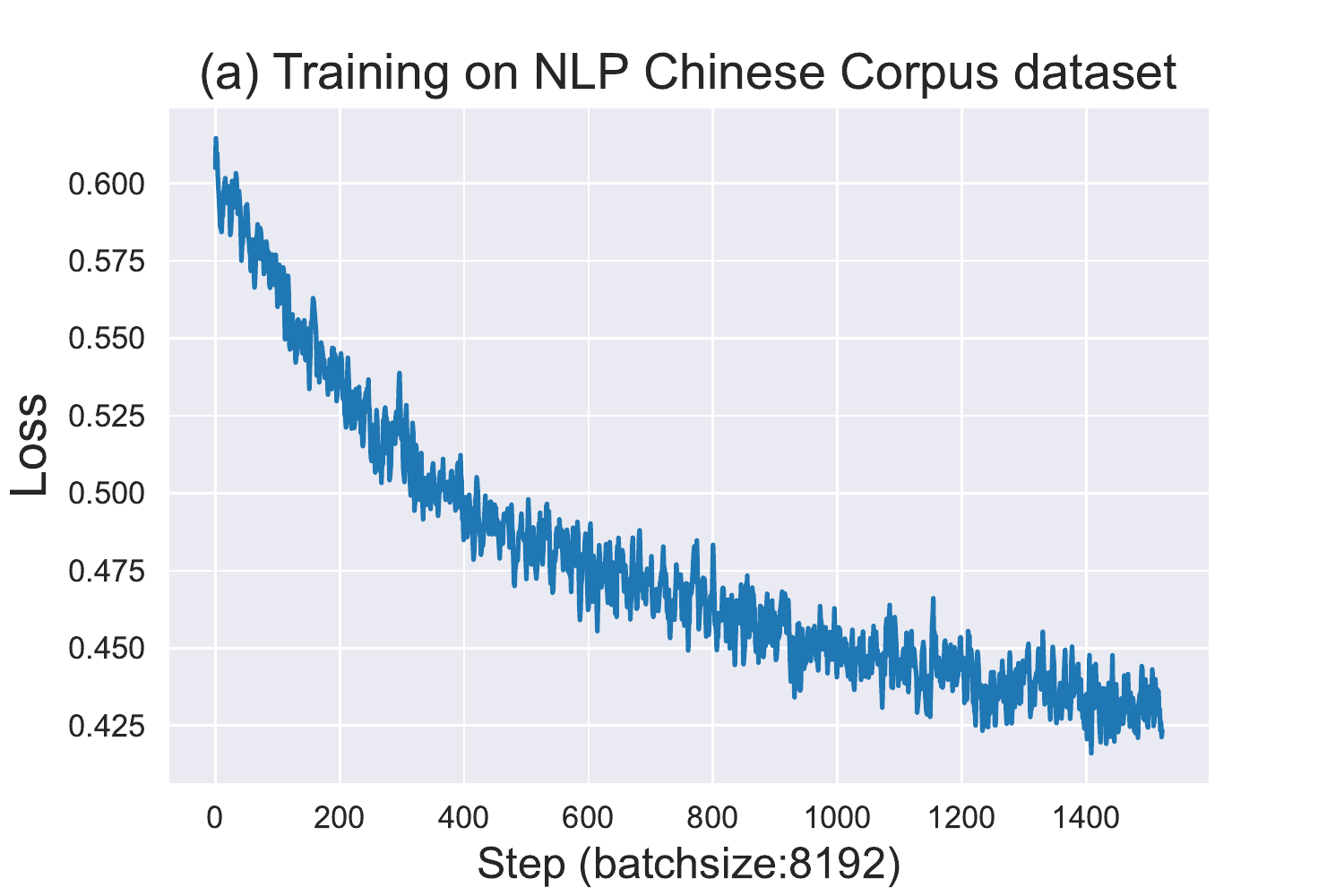}}
  \subfigure{\includegraphics[width=0.45\textwidth]{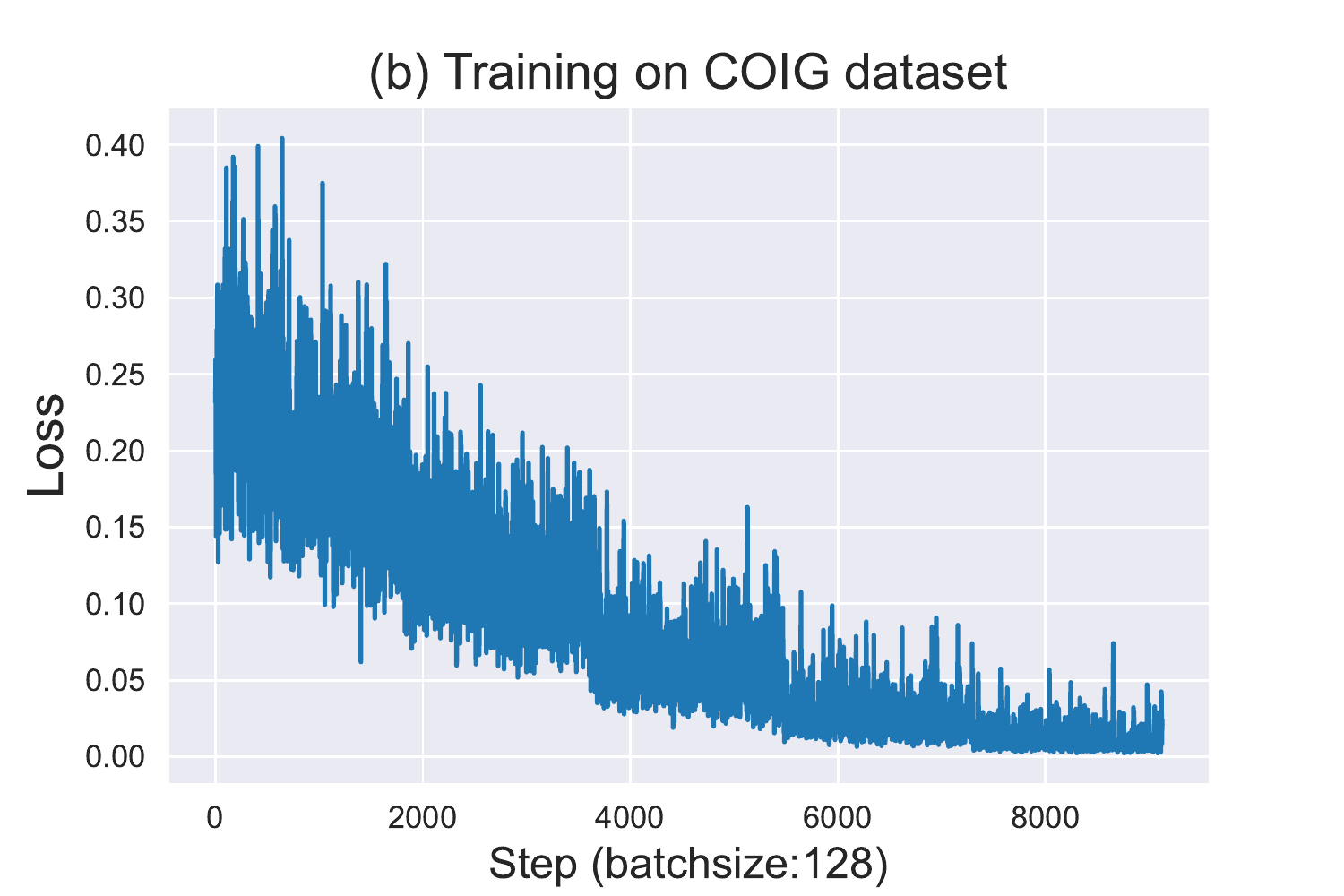}}
    \caption{Train steps versus losses on (a). Training on NLP Chinese Corpus dataset, and (b). Training on COIG dataset.}
    \label{fig:train_loss}
\end{figure}

%% file: contents/experiments.tex
\section{Experiments}
\label{sec:eval}

\begin{table*}[h]
\centering
\begin{tabular}{p{6cm} p{2cm} p{2cm} p{2cm}}

\hline
\textbf{Model} & \textbf{LogiQA-v2} & \textbf{$\mathbf{C^3}$-d} & \textbf{$\mathbf{C^3}$-m} \\
\hline
\hline
Linly-Chinese-LLaMA-7b-hf  & 25.91 & 32.28 & 34.52 \\
belle-llama-ext-7b \cite{ji2023exploring} & 26.41 & 29.52 & 28.87 \\
Panda-7B (ours) & 27.41 & 43.02 & 43.66 \\
Panda-Instruct-7B-3k steps (ours) & 26.22 & 39.05 & 42.11 \\
Panda-Instruct-7B-6k steps (ours) & 30.30 & 47.14 & 56.94 \\
\textbf{Panda-Instruct-7B-9k steps (ours)} & \textbf{31.93} & \textbf{47.30} & \textbf{57.04} \\

\hline
\end{tabular}
\caption{Experiment results for Panda-7B V.S. baselines on LogiQA-v2, $\text{C}^3$-d and $\text{C}^3$-m datasets.}
\label{tab:results}
\end{table*}

\subsection{Evaluation datasets}

We assessed the reasoning capabilities of our models using three publicly available reasoning benchmarks: LogiQA-v2 \cite{logiqav2}, which contains 8,678 QA instances; $\text{C}^3$ \cite{c3}, which contains 13k documents and their associated 19k Chinese multiple-choice free-form questions. For the $\text{C}^3$ dataset, we adopt $\text{C}^3$-Mixed, which contains non-dialogue documents of mixed genre, and $\text{C}^3$-Dialogue, of which the dialogue serves as the document. 

All three datasets provided us with a platform to evaluate the QA-reasoning capabilities of our language models. We have presented the relevant statistics of these datasets in Table \ref{tab:eval_data}.


\begin{table*}[h!]
\centering
\begin{tabular}{cccc}
\hline
Dataset & \# Samples & Format & Avg. length \\
\hline
\hline
LogiQA-v2 & 1594 & MCQA & 333 \\ 
$\text{C}^3$-d   & 1890 & Dialogue MCQA & 246 \\
$\text{C}^3$-m   & 2002 & Dialogue MCQA & 484 \\
\hline
\end{tabular}
\caption{The statistics of the evaluation datasets. We count the length of each sample as the tokenized sequence of context, question, and all options, using the sentence-piece tokenizer of Pre-trained LLaMA.}
\label{tab:eval_data}
\end{table*}

\subsection{Results}
\label{sec:results}

We show the experimental results in Table \ref{tab:results}. Specifically, we demonstrate the performance of Panda at different stages. 

\begin{itemize}
    \item Panda-7B: the model that is finetuned on Chinese-Wiki-2019, Chinese-News-2016, Chinese-Baike-2018, Chinese-Webtext-2019, and Translation-2019.
    \item Panda-7B-instruction-3k: Panda-7B + instruction tuning on COIG dataset for 3k steps.
    \item Panda-7B-instruction-6k: Panda-7B + instruction tuning on COIG dataset for 6k steps.
    \item Panda-7B-instruction-9k: Panda-7B + instruction tuning on COIG dataset for 9k steps.
\end{itemize}

From the results, we can observe that although a large amount of training effort was consumed in training our model on non-instruction conventional Chinese datasets, the performance of such a model is not desirable. In contrast, instruction-finetuning on COIG datasets provide a high boost to the performance of Panda. Specifically, with instruction tuning on COIG, which only takes up 4.2\% of our training samples, the performance of Panda increases from 27.41 to \textbf{31.93}, 43.02 to \textbf{47.30}, and 43.66 to \textbf{57.04} on LogiQA-v2, $\text{C}^3$-d and $\text{C}^3$-m respectively.


To provide a more comprehensive understanding of the training process, we present the training loss curves of Panda-7B on two datasets, namely the NLP Chinese Corpus dataset and the COIG dataset. Figure \ref{fig:train_loss} displays these curves. We observed that the training loss on the NLP Chinese Corpus dataset converges gradually until it reaches 0.425. We terminated the training process at approximately 1.5k steps as the model had trained on the entire dataset for one epoch. On the other hand, the training loss on the COIG dataset converged around 8k steps. We concluded training at 9k steps since the model had trained on the dataset for two epochs.

\subsection{Key findings}


\textbf{The key factor for achieving high performance in reasoning tasks is tuning on a diverse range of domains.} Our empirical experiments have shown that training on the NLP Chinese Corpus dataset alone is not enough to produce a high-performing model. To address this issue, we turned to the COIG dataset, which contains instruction data from a vast array of domains, including exam instructions, human value alignment instructions, Leetcode instructions, and more. As demonstrated in Section \ref{sec:results}, even using just 4.2\% of the COIG dataset dramatically improves our model's reasoning capability, particularly on the $\text{C}^3$-m dataset, with an impressive gain of 13.38.


\textbf{Mixing data indiscriminately does not lead to improved performance.} In an earlier attempt, we combined the NLP Chinese Corpus dataset with the COIG dataset and trained the entire dataset together. However, this approach did not yield better results and actually hindered the effectiveness of the COIG dataset. As a result, we only achieved a similar performance to the Panda-7B model without instruction tuning.

In a nutshell, a pipeline that incorporates abundant pretraining followed by instruction tuning on a small but diverse portion of data can lead to a highly effective Chinese language model.

%% file: contents/future.tex
\section{Upcoming Works}
The forthcoming objective involves the unveiling of more advanced models, namely Panda-13B, Panda-33B, and Panda-65B, which are characterized by their larger size and enhanced capabilities. In addition, the codes for enabling model parallel during training will be made publicly available, thereby benefiting the wider academic community. Furthermore, efforts will be directed towards the acquisition of additional training data, which will be utilized to improve the performance of both continual pre-training and instruction fine-tuning processes. Meanwhile, our attention will be focused on expanding the range of tasks and datasets included in the evaluation benchmark. Looking ahead, the ultimate goal is to incorporate more languages into our system, thereby further augmenting its versatility and adaptability.

\section{Conclusions}
This study focuses on the development and evaluation of Panda, an open-source Chinese instruction-following large language model. The performance of the model was assessed through experiments, which yielded results indicating that it outperforms existing open-source Chinese LLM initiatives and achieved state-of-the-art performance. The findings of this study may contribute to the improvement of open-source initiatives for large language models, as well as provide insight into effective model training strategies. By releasing training data, model checkpoints, and codes, we sincerely hope we can contribute to the democratization of AI.